\documentclass[12pt,a4paper]{article}

\usepackage[frontpage]{isr-report}
\usepackage[utf8]{inputenc}
\usepackage{amsmath}
\usepackage{amsfonts}
\usepackage[pdftex]{graphicx}

\begin{document}

\title{Online Event Segmentation in Active Perception using Adaptive Strong
  Anticipation}

\author{Bruno Nery \and Rodrigo Ventura}

\date{June 2010}

\trnumber{RT-701-10}

\support{This work was supported by the FCT (ISR/IST plurianual funding) through the PIDDAC Program funds.
    Partially funded with grant SFRH/BD/60853/2009, from Fundação para a Ciência e a
    Tecnologia}


\abstract{
Most cognitive architectures rely on discrete representation, both
in space (e.g., objects) and in time (e.g., events). However, a
robot interaction with the world is inherently continuous, both in
space and in time. 
The segmentation of the stream of perceptual inputs a robot receives
  into discrete and meaningful events poses as a challenge in bridging
  the gap between internal cognitive representations, and the external
  world.  Event Segmentation Theory, recently proposed in the context
  of cognitive systems research, sustains that humans segment time
  into events based on matching perceptual input with predictions.  In
  this work we propose a framework for online event segmentation,
  targeting robots endowed with active perception. Moreover, sensory
  processing systems have an intrinsic latency, resulting from many
  factors such as sampling rate, and computational processing, and
  which is seldom accounted for.  This framework is founded on the
  theory of dynamical systems synchronization, where the system
  considered includes both the robot and the world coupled (strong
  anticipation).  An adaption rule is used to perform simultaneous
  system identification and synchronization, and anticipating
  synchronization is employed to predict the short-term system
  evolution. This prediction allows for an appropriate control of the
  robot actuation. Event boundaries are detected once synchronization
  is lost (sudden increase of the prediction error).  An experimental
  proof of concept of the proposed framework is presented, together
  with some preliminary results corroborating the approach.}

%
 %% keywords/abstract page:
\maketitle
\tableofcontents
\pagebreak

\section{Introduction}
\label{sec:introduction}

The perception of a robot is grounded on the physical world. Its
sensors receive a continuous stream of information, as for instance
the light patterns hitting the CCD sensor of a video camera.
Cognitive representations, however, are often discrete, as in the case
of events and objects. Although the usage of digital computers demand
that all sensory information is discretized, this discretization is
commonly performed in fixed, not always adjustable, discretization
step (e.g, the frame rate and the pixel resolution of a video
camera). The detection of meaningful events from a stream of sensory
information is an important challenge, from the point of view of the
design of a cognitive architecture for robots, contributing to bridge
the gap between a continuous time world and discrete time,
event-based cognitive representations.

The segmentation of a continuous stream of information into events
is often overlooked, being commonly performed in an ad-hoc
manner, either recurring to threshold values over heuristic functions,
or fixed time triggers, for instance. But these methods are mostly
sensor modality dependent, as well as task specific. This work
addresses the problem of bridging the gap between the time continuous
stream of sensory/actuation information, and the discrete time
sequence of cognitive representations, proposing a modality and task
independent framework for event segmentation.

This problem is addressed using a biologically inspired
approach. Under this paradigm, our goal is not to faithfully model any
aspect of the human brain, but rather to employ findings from
neuroscience capable of providing guidance on how to engineer better
systems.

The Event Segmentation Theory (EST) provides a model of how the human
brain segments perception into a sequence of
events~\cite{zacks07,kurby07}. This model sustains that event
segmentation is based on the detection of prediction errors in the
sensory stream. Prediction is a commonplace mechanism found in many
brain systems. In particular, the human brain is permanently making
predictions and comparing them with the actual
outcome~\cite{schultz00}. Events are detected whenever a significant
disparity between prediction and outcome is encountered.  An event
segmentation mechanism can be built following this principle, but the
problem of how to make predictions about perceptions has to be
addressed first.

Dubois distinguishes between \emph{strong} and \emph{weak
  anticipation}~\cite{dubois03,stepp10}: the latter is based on an
explicit model of the world, where the physics is encoded
in analytical constructs, that can be mathematically solved given an
initial condition.  On the contrary, strong anticipation does not rely
on a model, but rather on the dynamical evolution of the interaction
of the agent with the world, seen as a single system. An example of
strong anticipation can be found on the behavior of an outfield
baseball player when catching a well-struck ball\footnote{Example
  from~\cite{stepp10}.}: weak anticipation of the ball landing
position requires modeling the physics of the ball, encoding the
initial state of the system (initial velocity, mass, friction
coefficient, etc), and then predicting the landing position by solving
the analytical model; in contrast, strong anticipation views the
outfielder and the ball as a single system with new dynamics, as the
outfielder moves itself driven by the projection of the ball on his
retina. Empirical evidence suggest that this is the way an human
outfield player functions~\cite{stepp10}.  In the context of robotics,
a model based approach to anticipation may be appropriate for passive
sensors, but when designing systems that actively engage in interactions
with the world, as in the case of active perception, the world
can no longer be modeled as an independent, self-contained system.

Stepp proposes an approach to strong anticipation based on the work
developed in the field of chaotic systems concerning synchronization
of dynamical systems~\cite{stepp10}. Consider two systems, denoted D
(drive) and R (response), connected by a unidirectional flow of
information from D to R. It is possible to design the system R such
that its dynamic evolution synchronizes with the one of D, regardless
of the initial condition of each system. One way of doing this is for
the R system to compare its state with the one of the D, and bias
its dynamics accordingly, \textit{i.e.,} system R is controlled by a
feedback loop, where the error results from this comparison.  More
interestingly, if this feedback loop contains a delay, system R is
capable, under certain conditions, to anticipate system
D~\cite{voss00}. Considering that system D includes both the robot and the
world, and system R to be a model internal to the robot, this
approach suggests an interesting mechanism to perform strong
anticipation of the dynamical evolution of the world-robot system.

One problem remains to be solved: how to design system~R? No system
model is assumed \textit{a priori}, since it depends on the coupling
involving the robot and the world.  A possible approach is to adapt
system~R during interaction.  A solution to the adaptation of response
systems in the context of dynamical systems synchronization has been
proposed by Chen~\cite{chen02}, where the convergence to the solution
has been proved using the Lyapunov stability theory.  This result does
not directly apply, however, to anticipating synchronization.

The contributions of this work are:
\begin{itemize}
\item An event segmentation method based on Stepp's strong
  anticipation concept~\cite{stepp10}, cast as an anticipating system
  synchronization framework;
\item The application of Chen's parameter identification
  method~\cite{chen02} to anticipating synchronization;
\item A proof-of-concept implementation of an architecture for event
  segmentation and active perception, employing these methods.
\end{itemize}

This report is organized as follows: after a short section surveying
related work, two sections on the theoretical background behind strong
anticipation and the adaptation method to learn the response system R
follow. Then, the proposed architecture for event segmentation is
described, followed by some experimental results of a proof of concept
implementation of these ideas. A section presenting some conclusions
and open questions closes the report.

\section{Related work}
\label{sec:related-work}

The problem of event segmentation has been studied in the
past. See~\cite{prem02} for a review of recent techniques for the
formation of event memories in robots.  Ramoni \textit{et al.}
proposed a method to cluster robot activities using Markov chain
models~\cite{ramoni00}.  In~\cite{guralnik99} a batch maximum likelihood
 estimator is used to fit a sequence of time-indexed models to
raw data. The incremental version of this algorithm is based on
thresholding the likelihood of the current model along time. The
spatio-temporal segmentation of video have been researched
in~\cite{wang94}, applying motion model clustering, and
in~\cite{dementhon02} using hierarchical clustering of the 3D
space-time video stream. Gesture segmentation and recognition has been
addressed in~\cite{kim07} employing hidden-Markov models (HMM).

\section{Strong anticipation}
\label{sec:strong-anticipation}

In~\cite{stepp10} strong anticipation is modeled using a dynamical
system synchronization framework. Consider two continuous dynamical
state vectors $x(t),y(t)\in\mathbb{R}^n$ with the following coupled
dynamics:
\begin{equation}
  \label{eq:1}
  \begin{split}
    \dot{x} &= f(x) \\
    \dot{y} &= f(y) + k(x-y_\tau)
  \end{split}
\end{equation}
where $y_\tau=y(t-\tau)$, \textit{i.e.,} a feedback loop with a
constant delay~$\tau$, and $k$ is a scalar gain. The first system is
called the \emph{drive} (D) while the second the \emph{response} (R).
This delayed feedback loop in the response system is a fundamental
aspect, and is responsible for the response system capability of anticipating
the trajectory of the drive.

This delayed feedback loop is neurophysiologically supported by the
discovery of forward models in the brain, which predict sensory
consequences of motor
commands~\cite{miall93,wolpert98,kawato99}. These models receive as
input a copy of the subject motor action, and produce a prediction of
future perceptions. For instance, when performing an arm movement,
these models predict the trajectory followed by the arm, as perceived
by the subject.  One important function of this mechanism is to
overcome the sensory processing latency in the brain, when the subject
is performing controlled, quick movements.

To understand how the response system can anticipate the drive,
consider that $\tau=0$ and that the systems are synchronized at time $t_0$,
\textit{i.e.,} $x(t_0)=y(t_0)$. Under these conditions, the systems
will remain synchronized, since $x-y_\tau=0$ and thus there is null
feedback in the response. In this case, the concatenated state
$z=(x,y)\in\mathbb{R}^{2n}$ evolves in the $x=y$ hyperplane, called the
\emph{synchronization manifold}~\cite{pecora97}. The response system
synchronizes with the drive if the error system with state $e=y-x$, also called the
\emph{transversal system}
\begin{equation}
  \label{eq:2}
  \dot{e} = f(y)-f(x)-k\,e
\end{equation}
is able to reject the perturbation $e$, driving it to zero. For
$f(y)\simeq f(x)$, system~(\ref{eq:2}) behaves like a first-order
system with an exponential decay to zero.  Anticipation is realized
once $\tau>0$, as synchronization implies $x(t)=y_\tau=y(t-\tau)$ and
thus $y(t)=x(t+\tau)$, meaning that the response anticipates the
driver. This is called \emph{anticipating
  synchronization}~\cite{voss00}, where $x=y_\tau$ defines the
\emph{anticipatory manifold}~\cite{voss01}.

Successful synchronization from an arbitrary initial condition is not
guaranteed in general (unless for simple cases), and strongly depends on
the values of $k$ and $\tau$. However, for any delay value $\tau$,
$e(t)=0$ is a fixed point of the transversal system~(\ref{eq:2}),
meaning that once synchronized, the system will remain so. Voss
conjectures that, if $e(t)=0$ is a stable fixed point for $\tau=0$,
then there is a $\tau_0>0$ such that, for any
$0<\tau<\tau_0$, the transversal system has a stable
fixed point at $e(t)=0$. This conjecture has been backed up by
numerical simulations~\cite{voss01}.

In general, for sufficiently small $\tau$, stability of the transversal
system can be expected. In the case of this work, since $\tau$ models the
delay of the perceptual system (e.g., the latency from a change in the
environment up to its detection by the computer vision algorithm),
this delay can be assumed smaller than the time scale of the events
being perceived.

\section{Adaptive synchronization}
\label{sec:adapt-synchr}

In the previous section it was assumed that the dynamics of the drive
and response systems are equal. If the drive system corresponds to the
world-robot coupled system, its dynamics is not known \textit{a
  priori}. One way of tackling this problem is to adapt the
response system, online, during synchronization.

Chen proposes in~\cite{chen02} an approach to adapt response systems
in the context of dynamical system synchronization. It does not account,
however, for a delayed feedback.

Consider that the drive system has the form
\begin{equation}
  \label{eq:3}
  \dot{x} = f(x) + F(x)\theta
\end{equation}
where $\theta\in\mathbb{R}^m$ is a vector of (constant) parameters,
$f(x)\in\mathbb{R}^n$ and $F(x)\in\mathbb{R}^{n \times m}$. The response system is
identical, except for the parameter vector that is unknown, and for the
synchronization feedback loop
\begin{equation}
  \label{eq:4}
  \dot{y} = f(y) + F(y)\alpha + U(y,x,t,\alpha)
\end{equation}
where $\alpha$ is the response parameter vector, and $U(y,x,t,\alpha)$
is called the controller of the response. Chen \textit{et al.} proved
in~\cite{chen02} that, under certain conditions,  not only the response
system synchronizes with the drive, but also that the response
parameters~$\alpha$ converge to the ones of the drive~$\theta$,
\textit{i.e.,}
\begin{equation}
  \label{eq:5}
  \lim_{t\rightarrow+\infty} ||\alpha(t)-\theta|| = 0.
\end{equation}
These conditions consist of the existence of a smooth controller
$U(y,x,t,\theta)$ and of a scalar (Lyapunov) function $V(e)$, where
$e=y-x$, such that:
\begin{enumerate}
\item $c_1||e||^2 \leq V(e) \leq c_2||e||^2$,
\item the derivative of $V(e)$ along the solution of the coupled system
  \begin{equation}
    \label{eq:7}
    \begin{split}
      \dot{x} &= f(x) + F(x)\theta \\
      \dot{y} &= f(y) + F(y)\theta + U(x,y,t,\theta)
    \end{split}
  \end{equation}
  satisfies $\dot{V}(e)\leq -W(e)$, and
\item the parameter vector $\alpha$ is adapted according to the
  learning rule
  \begin{equation}
    \label{eq:19}
    \dot{\alpha}(t) = -F^T(x) \left[ \nabla V(e) \right]^T
  \end{equation}
  for $\nabla V(e)$ denoting the gradient (row) vector of $V$ with respect
  to $e$,
\end{enumerate}
where $c_1$ and $c_2$ are two positive constants, $W(e)$ is a positive
definite function\footnote{$W(0)=0$ and $W(e)>0$ for any $e\neq0$.},
and $U(y,y,t,\theta)=0$.

This result has two important consequences: first, it proves
convergence, provided that the response system is capable of
synchronizing with the driver if $\alpha=\theta$ (\textit{i.e.,} if
the true parameters were known), and second, it provides a learning
law, in the form of the gradient of $\alpha$. However, in order to use
this result, one has to find a controller~$U$ and a function~$V$
satisfying the premises of the theorem. Chen shows that the controller
\begin{equation}
  \label{eq:8}
  U(y,x,t,\theta) = -e+f(x)-f(y)+\left[F(x)-F(y)\right]\theta
\end{equation}
and the Lyapunov function
\begin{equation}
  \label{eq:9}
  V(e) = \frac{1}{2} e^T e
\end{equation}
satisfy the premises for any~$F$ and~$f$.

The practical application of these results raises three practical
issues. One is the assumption that functions $F$ and $f$ are known,
meaning that one should have a prior knowledge of the structure of the
dynamics of the system. One can reverse this argument, stating that,
given functions $f$ and $F$ sufficiently generic, this method allows
the adaptation to any dynamical system that can be modeled
by~(\ref{eq:3}) for some parameter vector~$\theta$. Second, this
result was proved for continuous time systems. The discretization
of~$\dot{\alpha}$ raises the issue of the choice of a learning rate
(hidden in a proportionality constant of $V$, since the theorem is
invariant to a change of scale of this Lyapunov function). Finally,
the third issue concerns hidden state variables: if there is a state
variable that is hidden, \textit{i.e.,} the Lyapunov function $V(e)$
does not depend on its error, then this function is no longer positive
definite. This requires that all drive state variables have to be fed
to the response system controller. This is mostly
true\footnote{Occlusion of objects by others have to be accounted
  for.} once the state variables considered are all obtained from
perception (as in the case of the outfield baseball player example
above).

\section{Event segmentation}
\label{sec:event-detection}

\begin{figure}
  \centering
  \includegraphics[width=0.7\linewidth]{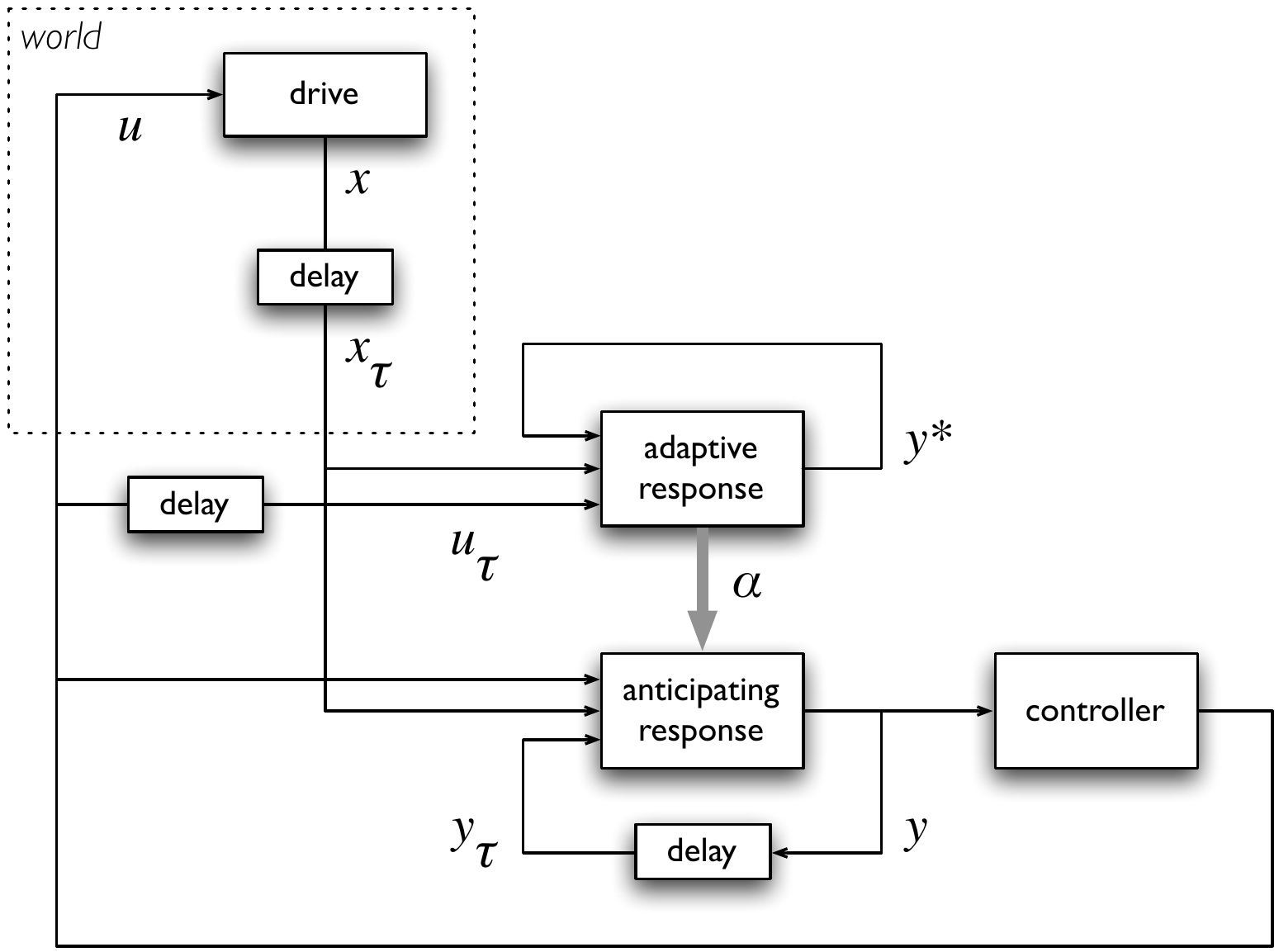}
  \caption{System architecture, consisting of the drive system and the
    perceptual delay (world), and the double response system formed by
    the adaptive and the anticipating responses. The anticipating
    response uses the parameters $\alpha$ obtained by the adaptive
    response. The control input $u$ is obtained by a controller fed
    with the anticipated state~$y$.}
  \label{fig:arch}
\end{figure}

The event segmentation framework we propose in this work, depicted in
Figure~\ref{fig:arch}, consists of a pair of response systems, one
performing adaptation (labeled \emph{adaptive response}), and the
other anticipation (labeled \emph{anticipating response}). The
adaptive response learns the parameter vector~$\alpha$ as described in
the Adaptive synchronization section, while anticipating response
performs anticipating synchronization as explained in the Strong
anticipation section. The robot-world coupled system is modeled by the
controlled drive system. Note that the access of the architecture to the world
state is subject to a delay, modeling for instance the latency of the
perceptual channel (image acquisition, processing, and tracking). The
controller computes the actuation vector~$u$ based on the anticipated
world state~$y$.

The drive system, together with the perceptual delay, is modeled by the
dynamical system
\begin{equation}
  \label{eq:10}
  \dot{x} = f(x) + F(x)\theta + u
\end{equation}
where $u$ is the control input, modeling the actuation of the robot in
the world. Shifting this equation by a delay of $\tau$ one obtains
\begin{equation}
  \label{eq:11}
  \dot{x}_\tau = f(x_\tau) + F(x_\tau)\theta + u_\tau
\end{equation}
where $u_\tau(t)=u(t-\tau)$. This model can be put in the form
of~(\ref{eq:3}) defining a time varying function
\begin{equation}
  \label{eq:12}
  f_\tau(x_\tau,t) = f(x_\tau)+u_\tau
\end{equation}
from which $\dot{x}_\tau=f_\tau(x_\tau,t)+F(x_\tau)\theta$.  The adaptive
response receives the delayed state $x_\tau$, together with the
delayed control input $u_\tau$
\begin{equation}
  \label{eq:13}
  \dot{y}^* = f(y^*) + F(y^*)\alpha + u_\tau + U(y^*,x_\tau,t,\alpha)
\end{equation}
Once $f_\tau(y^*,t)=f(y^*)+u_\tau$, this equation can be put in the form
of~(\ref{eq:4}). The anticipating response is described by
\begin{equation}
  \label{eq:14}
  \dot{y} = f(y) + F(y)\alpha + u + k(y_\tau - x_\tau)
\end{equation}
where $y_\tau=y(t-\tau)$ as before, and the parameter vector~$\alpha$
equals the one obtained by the adaptive response.  The anticipatory
 synchronization manifold is defined by $y_\tau=x_\tau$.  Thus,
$y=x$, meaning that the anticipating response is synchronized with the
drive system, which is the same to say that it is anticipating the delayed
perception~$x_\tau$. By shifting (\ref{eq:12}) in time one can get
$f_\tau(y,t+\tau)=f(y)+u$, allowing us to write~(\ref{eq:11})
and~(\ref{eq:14}) as
\begin{equation}
  \label{eq:10a}
  \begin{split}
    \dot{x}_\tau &= f_\tau(x_\tau,t) + F(x_\tau)\theta \\
    \dot{y} &= f_\tau(y,t+\tau) + F(y)\alpha + k(y_\tau - x_\tau) \\
  \end{split}
\end{equation}
thus matching~(\ref{eq:1}) (except for the time varying dynamics,
which do not affect the previous considerations on anticipating
synchronization) when $\alpha=\theta$.

According to the theory of Event Segmentation ~\cite{zacks07},
perceptual systems continuously make predictions about perceptual
input, and perceive event boundaries when transient errors in
prediction arise.  On the adaptive synchronization framework, the
Lyapunov function $V(e)$ defined in~(\ref{eq:9}), for $e=y^*-x_\tau$
provides a solid estimate of the prediction error.  Considering the
function values in a time window, we can associate the obtained samples
with a random variable with Normal distribution of mean $\mu_V$ and
variance $\sigma^2_V$. Under this assumption, the normalized metric
\begin{equation}
\label{eq:est}
b_V=\frac{V-\mu_V}{\sigma_V}
\end{equation}
is normally distributed with zero mean and unit variance.  When
$|b_V|$ exceeds a threshold $b_\textrm{event}$, an event boundary is
detected.  If $b_V$ is normally distributed with zero mean and unit
variance, the cumulative probability of the distribution tails for
$|b_V|>b_\textrm{event}$ is the probability of false positive
detection. Thus, $b_\textrm{event}$ should be sufficiently high so
that false positive detection is minimized, but low enough in order to
detect the prediction error increase due to a sudden change in the
dynamics of the system.

\section{Experimental results}
\label{sec:experimental-results}

\begin{figure}
\centering
\includegraphics[width=0.7\linewidth]{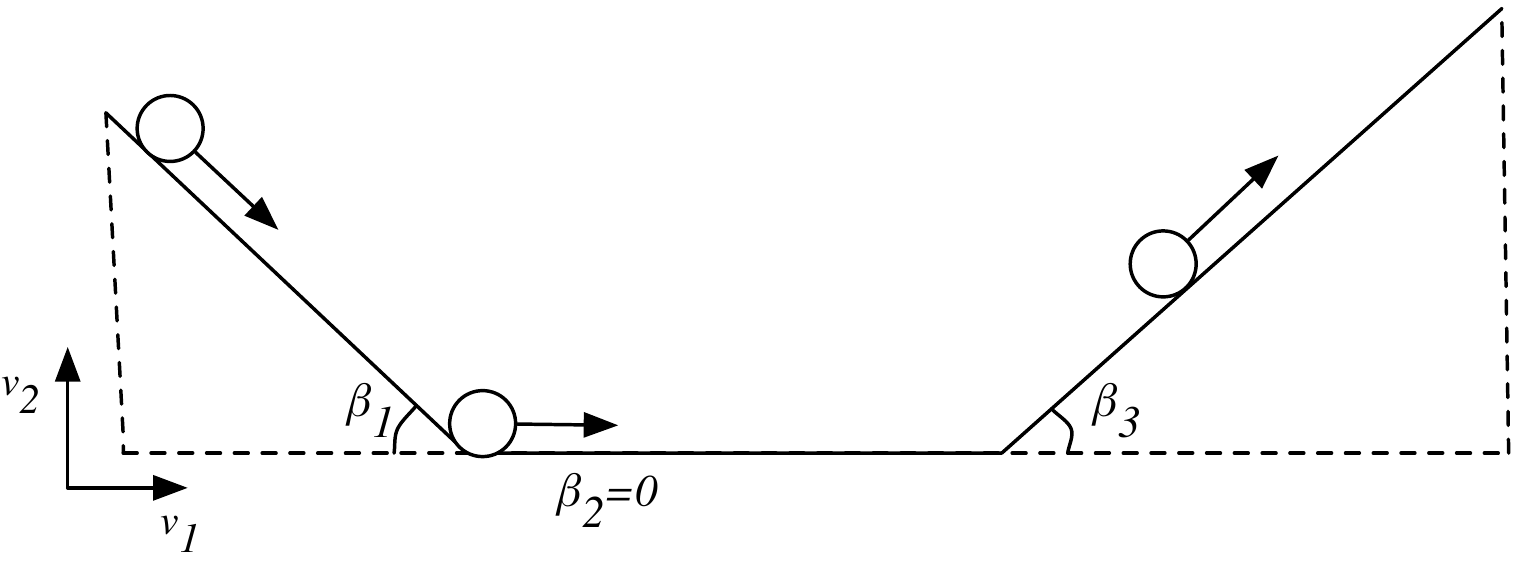}
\caption{Simulated scenario, where $\beta_1=\beta_3=\pi/12$.}
\label{fig:xr1_scenario}
\end{figure}

As a proof of concept for the ideas presented here, a simple scenario
was simulated: a ball rolling free on a series of inclined planes, with
different slopes, is observed by a robot camera which aims to follow it,
in order to center it on the image, as depicted in
Figure~\ref{fig:xr1_scenario}. The camera moves parallel to the plane,
for simplicity sake.

Denoting the ball coordinates by $v=[v_1\:v_2]^T$ and the camera
coordinates by $c=[c_1\:c_2]^T$, the ball projection $x=[x_1\:x_2]^T$
in the image plane is assumed orthographic: $x=v-c$. Assuming that
there is no ground friction, the dynamics of the ball is a double
integrator
\begin{equation}
  \label{eq:11a}
  \begin{split}
    \ddot{v}_1 &= -g\sin\beta\cos\beta \\
    \ddot{v}_2 &= -g\sin^2\beta \\
  \end{split}
\end{equation}
Considering that the camera support is frictionless and that its
movement is controlled in acceleration (\textit{i.e.,} force control),
the resulting drive system, in state space form, is given by
\begin{equation}
  \label{eq:6}
  \begin{split}
    \ddot{x}_1 &= - g\sin\beta\cos\beta - \ddot{c}_1 \\
    \ddot{x}_2 &= - g\sin^2\beta - \ddot{c}_2 \\
    \dot{x}_1 &= \dot{x}_1 \\
    \dot{x}_2 &= \dot{x}_2 \\
  \end{split}
\end{equation}
considering the state vector $x=[\dot{x}_1\:\dot{x}_2\:x_1\:x_2]^T$.
This system can be put in the form~(\ref{eq:10}) once
\begin{equation}
    \label{eq:15}
  \begin{split}
    f(x) &= \left[\begin{array}{c}
        0\\ 0\\ \dot{x}_1\\ \dot{x_2}
      \end{array}\right] \\
    \theta &= \left[\begin{array}{c}
        -g\sin\beta\cos\beta\\
        -g\sin^2\beta\\
      \end{array}\right]
    \end{split}
    \quad
    \begin{split}
      F(x) &= \left[\begin{array}{cc}
          1& 0\\
          0& 1\\
          0& 0\\
          0& 0\\
        \end{array}\right] \\
    u &= \left[\begin{array}{c}
        -\ddot{c}_1\\ -\ddot{c}_2\\ 0\\ 0
      \end{array}\right]
  \end{split}
\end{equation}
For this proof of concept, we set the response system to be
structurally identical, thus employing the same functions $f$ and $F$,
and control input $u$. The vector $\alpha=[\alpha_1\:\alpha_2]^T$ is
the parameter vector to be adapted according to Chen's learning
rule~(\ref{eq:19}).

When the anticipating response is synchronized with the drive, we have
$x=y$, and thus the dynamics of the anticipating response becomes
\begin{equation}
  \label{eq:20}
  \ddot{y} = \alpha - \ddot{c}.
\end{equation}
The camera motion controller considered has the form
\begin{equation}
  \label{eq:21}
  \ddot{c} = k_p y + k_d \dot{y} + \alpha
\end{equation}
where $k_p$ and $k_d$ are the proportional and the derivative gains of
the controller. Thus, the closed loop dynamics becomes
\begin{equation}
  \label{eq:22}
  \ddot{y} = -k_p y - k_d \dot{y}
\end{equation}
The design of the controller gains $k_p$ and $k_d$ can be
performed by pole placement (in the experiments we set $k_d^2=4k_p$,
yielding a smooth response with a double pole at $-k_d/2$).

The experiments were conducted after discretizing the above equations
using a simple approximation $\dot{z}(t)\simeq [z(t+T)-z(t)]/T$. The
sampling rate was 100Hz, $k_p=1$, $k_d=2$, $k=1$, and the Lyapunov
function used was~(\ref{eq:9}). The delay considered was $\tau=0.65s$
(65~samples). Event boundaries are detected using a 10-second window
and a $b_\mathrm{event}=3$. The system is initialized with the ball
starting on the top left position of the ramp, and as the ball
transverses the scenario there are two events, corresponding to the
two changes of the ramp slope. Each simulation takes~100s of simulated
time.

\begin{figure}
\centering
\includegraphics[viewport=130 243 490 540,width=0.7\linewidth]{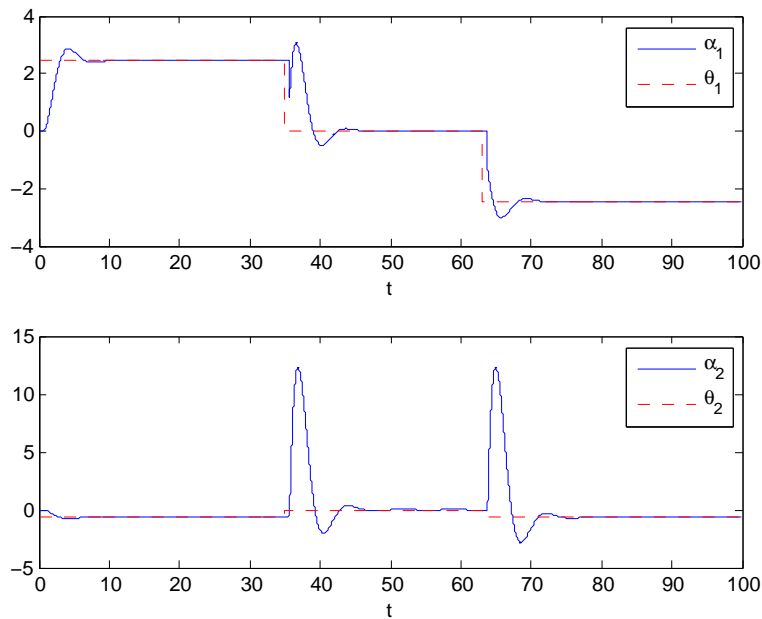}
\caption{Parameters $\alpha$ evolution (solid line) in comparison with
  the true values (dashed line).}
\label{fig:xr2_alpha}
\end{figure}

Figure \ref{fig:xr2_alpha} attests the performance of the adaptive
response system, in terms of the evolution of the parameters~$\alpha$,
compared with the ground truth ($\theta$, that changes with the
slope). As can be seen, the parameter vector $\alpha$ converges to the
true parameters $\theta$ after some time.

\def\myscale{0.9}

\begin{figure}
\centering
\includegraphics[viewport=130 243 490 540,width=0.7\linewidth]{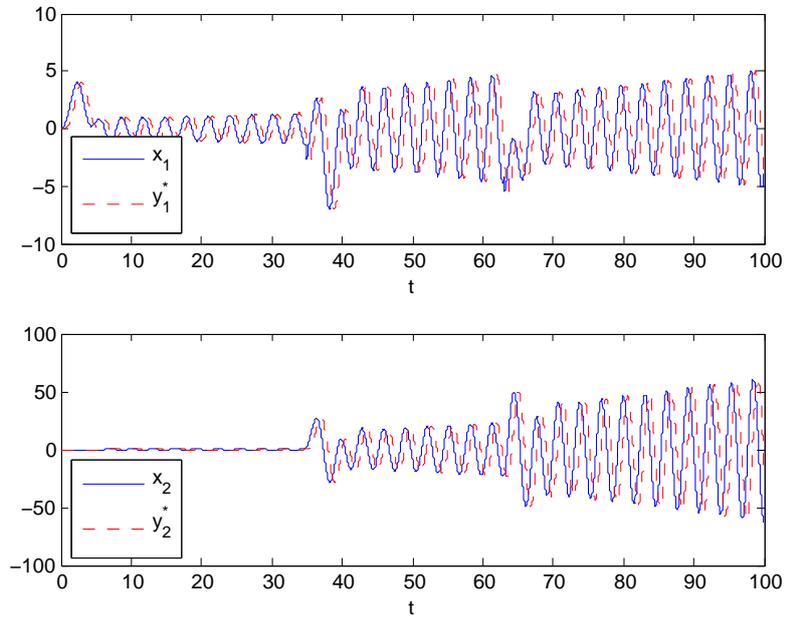}
\caption{System response without anticipation.}
\label{fig:xr3_xynoant}
\end{figure}

Figure \ref{fig:xr3_xynoant} shows the evolution of the ball position
in the camera without an anticipating response system, \textit{i.e.,}
the camera motion controller is fed by $y^*$ instead of $y$.  As
expected, the delay introduced by the latency of the perceptual
channel jeopardizes the control of the camera. Also, the adaptive
response follows the drive with a delay of $\tau$.

\begin{figure}
\centering
\includegraphics[viewport=130 243 490 540,width=0.7\linewidth]{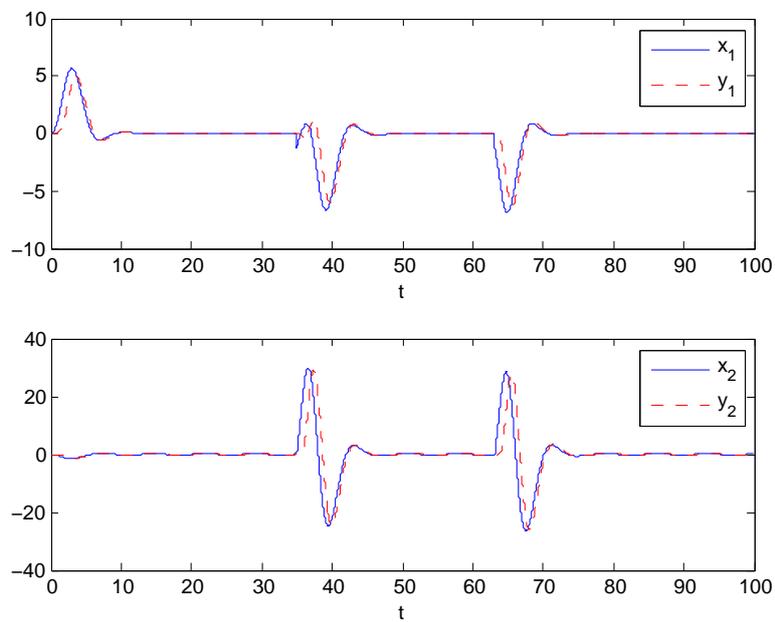}
\caption{System response using the full architecture.}
\label{fig:xr4_xyant}
\end{figure}

Figure \ref{fig:xr4_xyant} compares the ball position in the camera
with its anticipated response. In this case, both are synchronized,
since the ball coordinates in the image converge to zero (except for a
brief time after each slope change, while the adaptive system learns
the new parameters). Also, the anticipating response makes it possible
to control the drive system satisfactorily.

\begin{figure}
\centering
\includegraphics[viewport=130 243 490 540,width=0.7\linewidth]{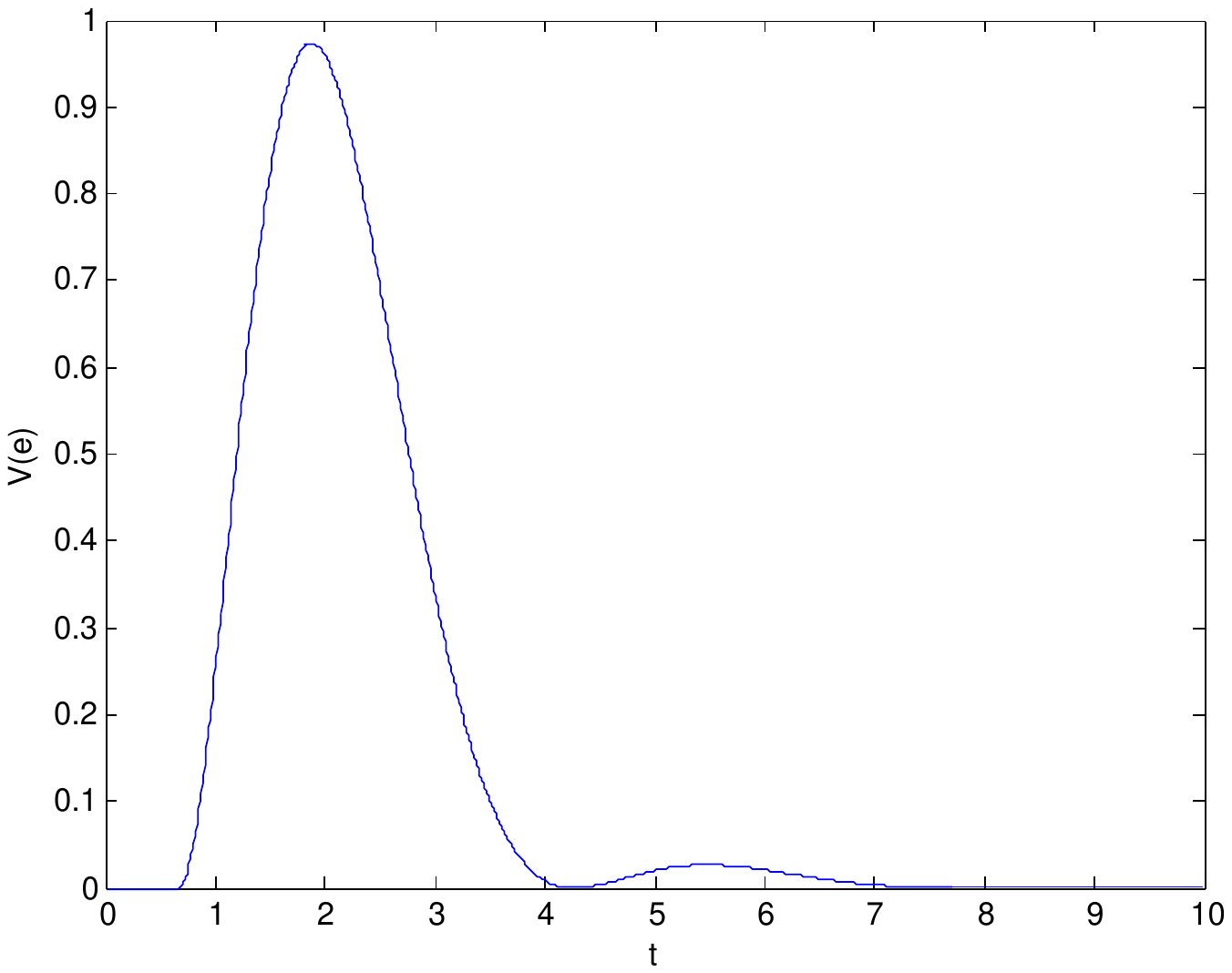}
\caption{Prediction error $V$ for the first 10 seconds of the simulation.}
\label{fig:xr5_Vant}
\end{figure}

Figure \ref{fig:xr5_Vant} pictures the evolution of the prediction error estimate
$V(e)$. Its value approaches zero as the drive and response system become
synchronized.

\begin{figure}
\centering
\includegraphics[viewport=130 243 490 540,width=0.7\linewidth]{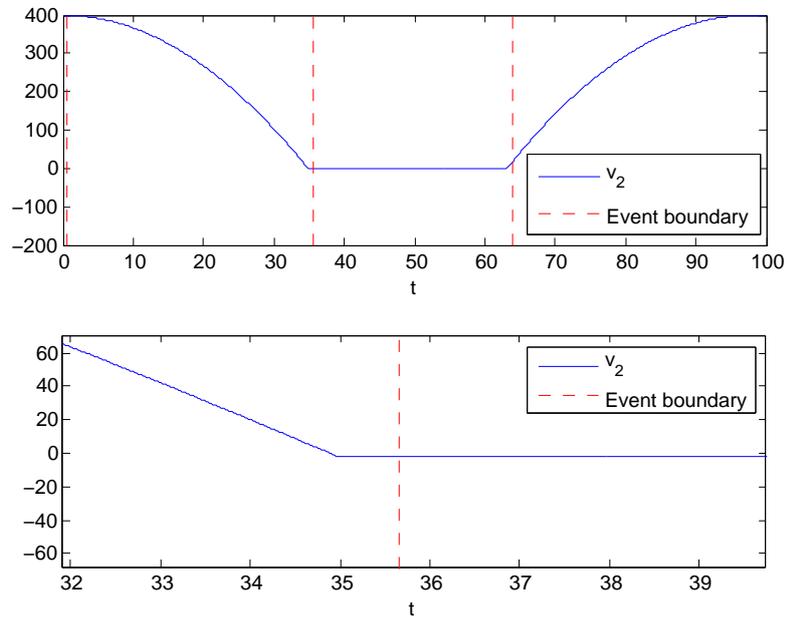}
\caption{The ball $v_2$ coordinate evolution along the experiment: top
plot shows the detected events, and the bottom plot a zoom around the
first detected event. The delay observed in this second plot
corresponds to the perceptual delay~$\tau$.}
\label{fig:xr6_est}
\end{figure}

Finally, Figure \ref{fig:xr6_est} shows the event segmentation results
obtained using the normalized metric~(\ref{eq:est}), with a window
of~10s. As expected, each change of plane is detected as an event
boundary by the framework. Interestingly, the peak of this metric, at
the event boundary, increases with the window size, without any loss
of temporal resolution.

These results show that the proposed system is capable of correctly
(1)~detecting the event boundaries that correspond to the change of ramp
slope by the ball, (2)~controlling the camera movement using anticipation,
and (3)~learning the correct system parameters.

\section{Conclusions and future work}
\label{sec:concl-future-work}

This report describes an event segmentation framework, targeting active
perception in robots, based on the concept of strong anticipation
proposed by Stepp \textit{et al.}  in~\cite{stepp10}. A dynamical
system synchronization paradigm is used as theoretical foundation of
the proposed architecture, where the robot-world coupled system is
identified using a parametric method for adaptation proposed by Chen
\textit{et al.} in~\cite{chen02}, and the actuation is performed
using anticipation. This anticipation accommodates for the net delay
of the perceptual channel. The capability of the architecture
to anticipate perception allows the robot to control its actuation
based on the prediction of the robot-world state, instead of relying on the
delayed perceptual data.

Having the described proof of concept experiments shown that the
proposed architecture behaves as expected, future work includes
scaling this approach to more complex domains. This involves tackling
the issues of the learning rate, which is hidden in the proportionality
constant of the Lyapunov function, used in the Chen's learning rule,
as well as the automatic design of the controller, given the adapted
parameters. Other open questions include dealing with hidden state
variables, as well as complex relations among objects (e.g., grasping,
occlusion, and so on).

\bibliographystyle{plain}
\bibliography{report}

\begin{thebibliography}{10}

\bibitem{chen02}
Shihua Chen and Jinhu L{\"u}.
\newblock Parameters identification and synchronization of chaotic systems
  based upon adaptive control.
\newblock {\em Physics Letters {A}}, 299:353--358, 2002.

\bibitem{dementhon02}
D.~DeMenthon.
\newblock Spatio-temporal segmentation of video by hierarchical mean shift
  analysis.
\newblock {\em Language}, 2, 2002.

\bibitem{dubois03}
Daniel~M. Dubois.
\newblock Mathematical foundations of discrete and functional systems with
  strong and weak anticipations.
\newblock In {\em Anticipatory Behavior in Adaptive Learning Systems}, Lecture
  Notes in Computer Science, pages 107--125. Springer, 2003.

\bibitem{guralnik99}
V.~Guralnik and J.~Srivastava.
\newblock Event detection from time series data.
\newblock In {\em Proceedings of the fifth ACM SIGKDD international conference
  on Knowledge discovery and data mining}, pages 33--42. {ACM}, 1999.

\bibitem{kawato99}
Mitsuo Kawato.
\newblock Internal models for motor control and trajectory planning.
\newblock {\em Current Opinion in Neurobiology}, 9(6):718--727, December 1999.

\bibitem{kim07}
Daehwan Kim, Jinyoung Song, and Daijin Kim.
\newblock Simultaneous gesture segmentation and recognition based on forward
  spotting accumulative hmms.
\newblock {\em Pattern Recognition}, 40(11):3012--3026, November 2007.

\bibitem{kurby07}
Christopher~A. Kurby and Jeffrey~M. Zacks.
\newblock Segmentation in the perception and memory of events.
\newblock {\em Trends in Cognitive Sciences}, 12(2):72--79, February 2008.

\bibitem{miall93}
R.C. Miall, D.~J. Weir, D.~M. Wolpert, and J.~F. Stein.
\newblock Is the cerebellum a smith predictor?
\newblock {\em Journal of Motor Behavior}, 25(3):203--216, 1993.

\bibitem{pecora97}
Louis~M. Pecora, Thomas~L. Carroll, Gregg~A. Johnson, and Douglas~J. Mar.
\newblock Fundamentals of synchronization in chaotic systems, concepts, and
  applications.
\newblock {\em Chaos}, 7(4):520--543, 1997.

\bibitem{prem02}
Erich Prem, Erik H{\"o}rtnagl, and Georg Dorffner.
\newblock Growing event memories for autonomous robots.
\newblock In {\em Proceedings of the Workshop On Growing Artifacts that Live:
  Basic Principles and Future Trends}, 2002.

\bibitem{ramoni00}
Marco Ramoni, Paola Sebastiani, and Paul Cohen.
\newblock Unsupervised clustering of robot activities: a bayesian approach.
\newblock In {\em Proceedings of the fourth international conference on
  Autonomous agents ({AGENTS'00})}, pages 134--135, 2000.

\bibitem{schultz00}
Wolfram Schultz and Anthony Dickinson.
\newblock Neuronal coding of prediction errors.
\newblock {\em Annual Review of Neuroscience}, 23:473--500, 2000.

\bibitem{stepp10}
N.~Stepp and M.T. Turvey.
\newblock On strong anticipation.
\newblock {\em Cognitive Systems Research}, 11:148--164, 2010.

\bibitem{voss00}
Henning~U. Voss.
\newblock Anticipating chaotic synchronization.
\newblock {\em Physical review {E}}, 61(5):5115--5119, 2000.

\bibitem{voss01}
Henning~U. Voss.
\newblock Dynamic long-term anticipation of chaotic states.
\newblock {\em Physical Review Letters}, 87(1):14102, July 2001.

\bibitem{wang94}
J.Y.A. Wang and E.H. Adelson.
\newblock Spatio-temporal segmentation of video data.
\newblock In {\em {SPIE} Proceedings Image and Video Processing II}, volume
  2182, pages 120--131, 1994.

\bibitem{wolpert98}
Daniel~M. Wolpert, R.~Chris Miallb, and Mitsuo Kawato.
\newblock Internal models in the cerebellum.
\newblock {\em Trends in Cognitive Sciences}, 2(9):338--347, 1998.

\bibitem{zacks07}
Jeffrey~M. Zacks, Nicole~K. Speer, Khena~M. Swallow, Todd~S. Braver, and
  Jeremy~R. Reynolds.
\newblock Event perception: A mind–brain perspective.
\newblock {\em Psychological Bulletin}, 133(2):273--293, 2007.

\end{thebibliography}

\end{document}